\pdfoutput=1

\documentclass[11pt]{article}

\usepackage[]{ACL2023}

\usepackage{times}
\usepackage{latexsym}
\usepackage{multirow}
\usepackage{tabularx}
\usepackage{graphicx}
\usepackage{todonotes}
\usepackage{amssymb} 
\usepackage{booktabs}
\usepackage{comment}
\newcommand{\highlight}[1]{\colorbox{blue!10}{#1}}
\newcommand{\unhighlight}[1]{\colorbox{red!10}{#1}}
\usepackage{hyperref}
\usepackage{scrextend}
\usepackage{tablefootnote}
\usepackage{longtable}
\usepackage{times}
\usepackage{latexsym}
\usepackage{amsmath}

\usepackage{array}
\usepackage[T1]{fontenc}

\usepackage[utf8]{inputenc}

\usepackage{microtype}
\newcolumntype{L}{>{\centering\arraybackslash}m{2.5cm}}
\newcolumntype{S}{>{\centering\arraybackslash}m{1.5cm}}

\newif\ifcomments
\commentstrue
\ifcomments
    \providecommand\kb[1]{\textcolor{teal}{[KB: #1]}}
    \providecommand\partha[1]{\textcolor{olive}{[PT: #1]}}
    \providecommand\shachi[1]{\textcolor{red}{[SD: #1]}}
\else
    \providecommand{\kb}[1]{}
    \providecommand{\partha}[1]{}
    \providecommand{\shachi}[1]{}
\fi

\usepackage{inconsolata}

\title{Parameter-Efficient Finetuning for Robust Continual Multilingual Learning}

\author{Kartikeya Badola \hskip 1em  
Shachi Dave  \hskip 1em
Partha Talukdar \\
        Google Research India \\  
   \texttt{\{kbadola, shachi, partha\}@google.com}}

\begin{document}
\maketitle
\begin{abstract}
We introduce and study the problem of \textbf{Continual Multilingual Learning} (CML) where a previously trained multilingual model is periodically updated using new data arriving in stages. If the new data is present only in a subset of languages, we find that the resulting model shows improved performance only on the languages included in the latest update (and a few closely related languages) while its performance on all the remaining languages degrade significantly.  We address this challenge by proposing \textbf{LAFT-URIEL}, a parameter-efficient finetuning strategy which aims to increase the number of languages on which the model improves after an update, while reducing the magnitude of loss in performance for the remaining languages. LAFT-URIEL uses linguistic knowledge to balance overfitting and knowledge sharing across languages, allowing for an additional 25\% of task languages to see an improvement in performance after an update, while also reducing the average magnitude of losses on the remaining languages by 78\% relative.
\end{abstract}

\section{Introduction}
\label{sec:intro}

A learning-based NLP model may need to be periodically updated for a variety of reasons, e.g., to incorporate newly available training data, adapt to data shifts, etc. Continual learning \citep{thrun1995lifelong,kirkpatrick2017overcoming} and Online learning \citep{shalev2012online} are paradigms where a model is sequentially trained on packets of new training data, without having access to old training data. In such settings, the goal is to ensure that the model is able to incorporate incremental knowledge from the new data without forgetting the knowledge obtained from prior training.

As multilingual NLP grows in prominence, the underlying models used for multilingual tasks are increasingly being developed as a single deep neural network trained on data from all supported languages \citep{devlin-etal-2019-bert, conneau-etal-2020-unsupervised, xue-etal-2021-mt5}. Having a shared multilingual model instead of one model per language allows one to reduce the number of models to train and maintain for the downstream task, improve  performance on lower-resource languages due to cross-lingual sharing of knowledge, and improve inference on code-mixed inputs. Just like monolingual models, multilingual NLP models also need to be regularly updated, thereby making them suitable for application of continual learning strategies.

However, continual learning of multilingual models  involves additional challenges due to involvement of multiple languages. For instance, during any update, the new training data may cover only a small subset of the languages, which may negatively impact the performance on languages not represented in this new data. This scenario is often true for multilingual models deployed in production settings. 
In spite of its importance and real-world significance, the problem of \textbf{Continual Multilingual Learning (CML)} has not been much explored. We fill this gap in this paper.

In the CML setting, a single-task multilingual model 
needs to be continually updated with additional training data from a subset of the supported languages arriving in stages, while keeping the model capacity fixed and without relying on any data from previous training steps. 
\begin{figure*}[t!]
\centering
\includegraphics[width=9.2cm]{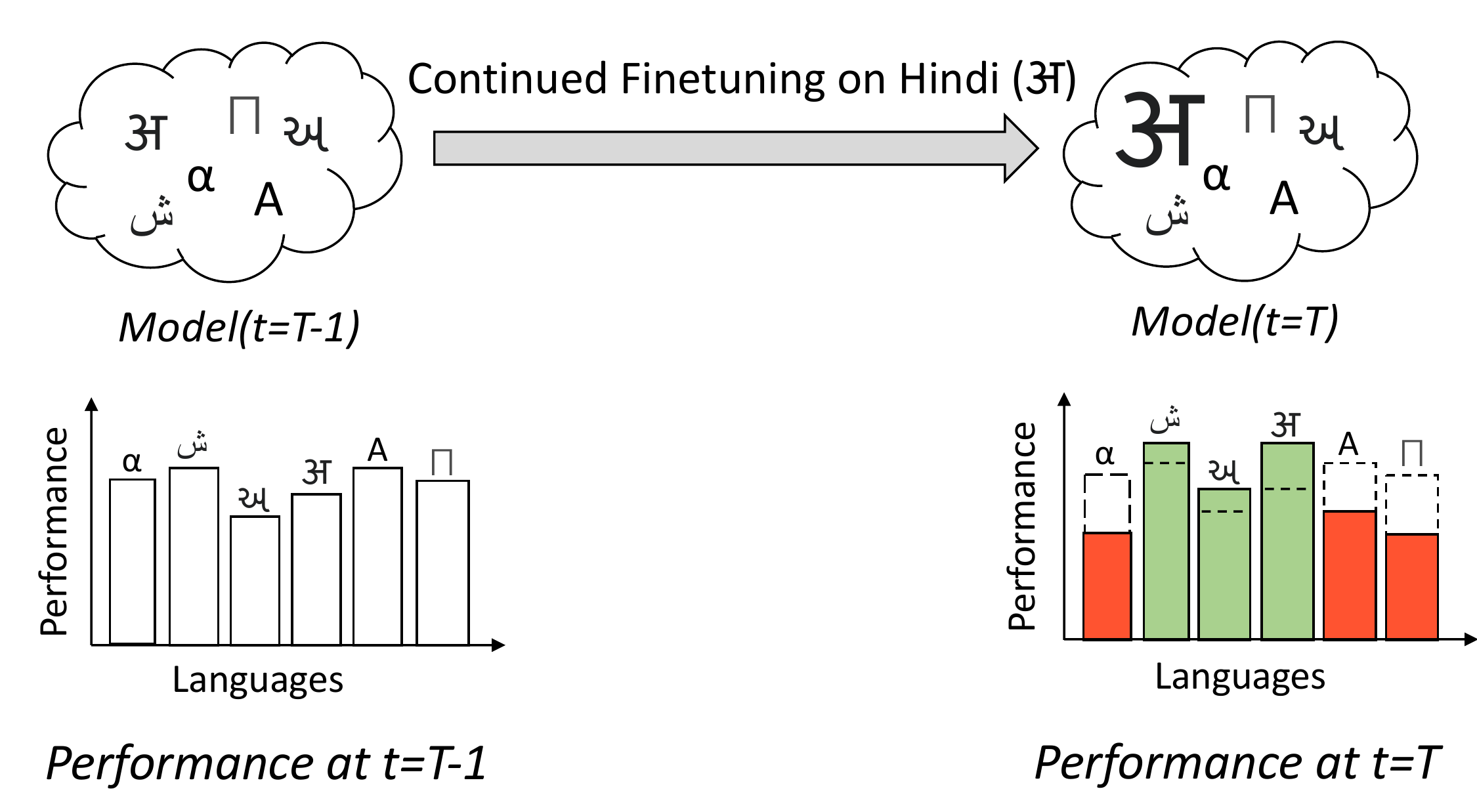}
\caption{\label{fig:experimentation_setup}
In the Continual Multilingual Learning (CML) setup, a previously trained multilingual model at time $\mathrm{T - 1}$ is further finetuned on new data (for the same task) coming from a subset of seen languages (only Hindi in this example), resulting in the updated model at time $\mathrm{T}$. We observe that the additional training results in improved performance on the new data languages (Hindi) and a few closely related languages (due to positive transfer), while negatively affecting the remaining languages. This paper proposes strategies to maximize positive transfer while minimizing negative impact on the remaining languages during CML. 
}
\end{figure*}
Given a shared multilingual model, the goal of updating it on new data for the same task would be to (1) improve the model performance across most, if not all, languages and (2) ensure that none of the languages incur a significant loss in performance. The second scenario may occur if the new training data is highly skewed towards a subset of languages, making it easier for the model to overfit on the language specificities of the new data while forgetting the same for languages not represented in this update. In our study, we find that balancing the two goals is non-trivial and the model incurs significant losses across a subset of languages if it is finetuned on the new data in an unconstrained manner. We study this phenomenon over four tasks and find the same non-ideal behaviour across all experiments for the baseline finetuning strategy. The CML setup is closest in spirit to 
\citet{m2022cross}, where a multilingual model is trained from \emph{scratch} with additional model parameters added during updates. In contrast, CML builds on top of an \emph{existing} multilingual model while keeping model capacity \emph{fixed}. 

We start with the intuition that constraining the number of trainable parameters in the network would help control for losses due to language-specific forgetting. 
We operationalize this through different parameter-efficient finetuning strategies, namely Adapters \citep{houlsby2019parameter} and Composable Sparse-finetuning \citep{ansell-etal-2022-composable}. We find that such methods provide a middleground by allowing limited cross-lingual sharing of knowledge while reducing model's tendency to overspecialize on the languages of the new data.

With this initial promise, we develop \textbf{LAFT-URIEL}, a novel finetuning strategy which uses Adapters and URIEL language similarity metrics \citep{littell-etal-2017-uriel} to balance the trade-off between encouraging positive cross-lingual transfer and discouraging language-specific forgetting. Our contributions are as follows:
%
\begin{enumerate}
    \item We introduce and study Continual Multilingual Learning (CML) where a multilingual model is periodically updated with batches of new data from a subset of the languages covered. This is an important but unexplored problem of practical significance. 
    \item In the CML setup, we show that a model may suffer from drastic language-specific losses if the new training data is skewed towards a subset of languages, thus making the resulting model unfit for multilingual downstream applications.
    \item We propose LAFT-URIEL, a novel finetuning strategy which uses Adapters and syntactic language similarity to maximize positive transfer during CML, while minimizing negative impact across languages.  
\end{enumerate}
We present the CML setup in Figure \ref{fig:experimentation_setup}.
\section{Problem Setup}
\label{sec:problem_setup}

We consider a setting where a trained, task-specific multilingual model, which we will call as the \textit{deployed model}, is further finetuned on new finetuning data for the same task, to give us the \textit{updated model}. To ensure the best possible multilingual performance, we will assume that the deployed model has been previously trained on data from all supported languages for the task (say $N_L$ in number).

In an update, the deployed model will be finetuned on new task-specific data, to give us the updated model. In the real world, as one may have no control over how the new data is distributed across languages, we will assume the worst case scenario for our setup (i.e., maximum skew) where the new data is only present in one of the $N_L$ languages.
\\
We divide the entire setup into two stages:

\vspace{1ex}
\noindent \textbf{\emph{1. Inception stage}} where we setup the first deployed model by training a transformer model (initialized by pre-trained mBERT \citep{devlin-etal-2019-bert} or XLM-RoBERTa \citep{conneau-etal-2020-unsupervised} checkpoints) on task data in all $N_L$ languages.

\vspace{1ex}
\noindent \textbf{\textit{2. Continuation stage}} where we further finetune the deployed model on the new finetuning data (in one of the $N_L$ languages) to give us the updated model. There can be multiple continuation stages that are sequentially performed one after another.\footnote{The dev and test sets are kept the same across all finetuning stages and covers all languages for the given task.}
\\
Formally, we define our setup using the following notations. For inception stage: 
\begin{gather}
    \mathrm{model(t=0^-)}\xrightarrow{\{l_1, l_2, ..., l_{N_L}\}}\mathrm{model(t=0)}
\end{gather}
where $\mathrm{model(t=0^-)}$ is the untrained model, $\mathrm{model(t=0)}$ is the first deployed model and $\mathrm{model(t=T)}$ is model after $\mathrm{T}$ continuation stages. $\xrightarrow{\{l_{i_1}, l_{i_2}, ..., l_{i_k}\}}$ denotes finetuning using data from languages $\{l_{i_1}, l_{i_2}, ..., l_{i_k}\}$. 
The $\mathrm{T^{th}}$ continuation stage can be written as:
\begin{gather}
\mathrm{model(t=T-1)} \xrightarrow{\{l_i\}} \mathrm{model(t=T)}
\end{gather}
where $i\in \{1,2,...,N_L\}$.
\\ \\
To compare different finetuning strategies, we focus on the $\mathrm{t=0}$ and $\mathrm{t=1}$ transition and subsequently study the effects of multiple sequential continuation stages on the model using our proposed strategy. For a task in $N_L$ languages, there can be $N_L$ different continuation stages to transition from the deployed model at $\mathrm{t=0}$ to the updated model at $\mathrm{t=1}$ (since the continuation stage data can come from any one of the $N_L$ task languages). We consider all such cases. Training data for each finetuning stage is created by partitioning the full training data into equal parts, independently across all languages.

On a fixed test set, we expect the language-wise performance of the deployed model and the updated model to differ due to multiple reasons: (1) additional task-learning using new task data (2) catastrophic forgetting of language-specific knowledge (3) positive or negative cross-lingual transfer. Ideally, we would want the performance delta to be positive or neutral across each language. Hence, the goal is to devise strategies that encourage language-agnostic task learning and positive cross-lingual transfer while inhibiting catastrophic forgetting.

We select four representative tasks from three families: token-level, sentence-level and seq2seq; in order to ensure that our methodology is broadly applicable. These are: PAN-X (aka WikiANN) \citep{pan-etal-2017-cross} for NER tagging, Universal Dependencies v2.5 for POS tagging (UDPOS) \citep{nivre2018universal}, MTOP \citep{li-etal-2021-mtop} for domain classification and semantic parsing (NSP). More details for each task is provided in Table \ref{table:data_stats} and Appendix \ref{sec:appendix_dataset}. For PAN-X and UDPOS, the language selection is done based on the pre-trained weights available for Lottery Ticket Sparse Fine-Tuning (\S \ref{subsec:background}). The resulting set of languages across tasks offer a diverse mix of typologies, language families and geographic location of prominence. 

\begin{table}[t]
\centering
\small{\begin{tabular}{@{}ccc@{}}
\toprule
  \textbf{Dataset} & \textbf{Task Type} & \textbf{Languages} \\ 
 \hline
 PAN-X & Token level & en, hi, bn, zh, ta, ja, ar  \\ 
 UDPOS &  Token level & en, hi, ja, ta, zh, ar \\
 MTOP Class. & Sentence level & en, de, es, fr, hi, th \\
 MTOP NSP &  Seq2seq & en, de, es, fr, hi, th \\ \bottomrule
\end{tabular}}
\caption{Tasks studied in the CML setup. We use the ISO 639-1 language codes to denote languages and present the mapping in Appendix \ref{sec:appendix_dataset}.}
\label{table:data_stats}
\end{table}

Each experiment is repeated three times by varying the random seed. The seed also varies the examples selected for constructing finetuning sets of different stages.

\section{Baseline Finetuning Strategy and Metrics}
\label{sec:baseline}
The baseline finetuning strategy in our setup would be finetuning while keeping all the parameters of the model trainable during a continuation stage. We call this the \textit{Full Finetuning} (\textbf{FFT}) baseline.

We anticipate that continued finetuning on new task data (which is skewed towards a particular language) would cause non-uniform changes in language-wise performance on a fixed test set. In particular, we expect performance gains on the language seen during a continuation stage and losses across some subset of the remaining languages. 

To compare different finetuning strategies, we measure the percentage change in language-wise performance after the continuation stage\footnote{We use percentage change as opposed to magnitude change since negative change in performance on a language which had poor base performance should be penalized more than the same for a language with high base performance.}.

For the updated model to be fit for deployment (e.g., in a production setting), it is necessary to ensure that the performance drop on any language is not too high. Also, given the shared multilingual model, an ideal strategy should be able to spread the gains in performance across most if not all languages. To this end, we construct the following metrics to compare different strategies in our setup:

\vspace{0.5ex}
\noindent
$\mathbf{AvgPercentLoss}$: \textit{Average magnitude of percentage loss after continuation}. Calculated by averaging the absolute percentage change in performance over all languages which suffered a loss in performance. For an ideal model, this should be 0.

\vspace{0.5ex}
\noindent
$\mathbf{NumImprovedLangs}$:
\textit{Average number of languages with a positive change in performance after a continuation stage}. For an ideal model, this should be the count of all supported languages for a given task, $N_L$.
\\
Since there can be $N_L$ different continuation stages (where the new data is only present in one of $N_L$ languages for the task), we report the average of the above two metrics across all such transitions.

\vspace{0.5ex}
\noindent
\textbf{Additional constraints}: After a continuation stage, we would at the very least expect that, (1) the sum of gains are higher than the magnitude of the sum of losses and (2) the magnitude of maximum gain are higher than the magnitude of the maximum loss in performance. A finetuning strategy which is unable to obey these constraints can simply be declared as unfit for our setup. We therefore compute an average of $\mathrm{sum(gains)/abs(sum(losses))}$ ($\mathbf{SumRatio}$) and $\mathrm{max(gains)/abs(max(losses))}$ ($\mathbf{MaxRatio}$) across the different continuation stages and check whether the two values are $\geq1$.
\section{Parameter-efficient Finetuning for Continual Multilingual Learning}
\label{sec:param_efficient}
We propose the use of parameter-efficient finetuning methods to build improved finetuning strategies in our setup. The benefits of using these methods would be two-fold. Firstly, such methods should allow one to constrain the changes being made in the model, which should help in controlling the losses due to forgetting. Secondly, these methods can be used to decompose task learning into language-specific and language-agnostic parts. This property can be used to update the model in a language-agnostic fashion which should help in spreading gains across languages.

The following subsection will give an overview of the methods we intend to use to build improved finetuning strategies for the task.
\subsection{Methodologies}
\label{subsec:background}
\vspace{0.5ex}
\noindent
\textbf{Lottery Ticket Sparse Fine-Tuning (LT-SFT)} \citep{ansell-etal-2022-composable}: proposes to keep only a subset of parameters trainable during finetuning. This allows one to learn a sparse vector of differences (\textit{update matrix}) with respect to the base model. Update matrices for different sub-tasks can be composed together by simply summing up the diffs.

Using the above compositionality property, one can build a pipeline to decompose multilingual task learning into task-specific and language-specific parts. For the language-specific part, we use the pre-trained sparse update matrices for each language, obtained by finetuning on language-specific data for masked language modelling.

Given finetuning data for a task, the task-specific sparse updates are learnt by first applying the language-specific update matrix for the language of the training example and then performing gradient descent on this sparsely modified model. The learned vector of differences can be assumed to be language-agnostic since the model already has language-specific knowledge from the update matrix applied before the forward pass. For multilingual finetuning (e.g., during inception stage), we follow \textit{multi-source} training where data batches are constructed per language and uniformly sampled across languages throughout finetuning. We use LT-SFT to build a stronger baseline for the task, which we will call \emph{\textbf{SFT} (sparse finetuning)}.

\vspace{0.5ex}
\noindent
\textbf{Adapters} \citep{houlsby2019parameter}: Adapters are trainable modules that are inserted in the layers of a transformer network. During finetuning, usually only the adapter modules are kept trainable and these constitute to about 5-10\% of the parameters of the network.

In our work, we use adapters to split the model into language-agnostic and language-specific parts and propose a finetuning strategy called \emph{\textbf{LAFT} (language-specific adapter finetuning)}.

\vspace{0.5ex}
\noindent
\textbf{URIEL vectors} \citep{littell-etal-2017-uriel}: We propose to use URIEL vectors to estimate whether language-specific learning for a given language would be useful for another language by computing the URIEL syntactic distance between the two languages. The syntactic distance is computed as the cosine similarity between the syntactic vectors of any two languages obtained from the URIEL database. 

Prior works such as MAD-G \citep{ansell-etal-2021-mad-g} have used URIEL vectors in conjunction with parameter-efficient finetuning methods for generating adapter modules for unseen languages. The generated adapters sometimes perform slightly worse than vanilla adapters on the seen set of languages depending upon the task. We hence stick with vanilla adapters and propose our novel finetuning strategy, \emph{\textbf{LAFT-URIEL}}, for integrating knowledge from the URIEL vectors.

\subsection{Proposed Finetuning Strategies}
\label{subsec:improved_pipelines}
We build our finetuning strategies using the methodologies described in \S \ref{subsec:background} and describe the inception and continuation stage for each case. In each strategy our goal is to a) make minimal changes to the shared parameters of the deployed model and b) ensure that such changes are language-agnostic. This should help in spreading the performance gains while also minimizing losses.

\vspace{0.5ex}
\noindent
\emph{\textbf{Sparse Finetuning (SFT)}}
For the inception stage we follow the standard \textit{multi-source} training (\S \ref{subsec:background}) using the pre-trained language-specific sparse update matrices. In this stage, the base model is sparsely trainable and the classifier (or the decoder for the seq2seq task) is fully trainable\footnote{This gives us a better performing deployed model compared to when the decoder is kept sparsely trainable too.}.

During the continuation stage, we sparsely update the entire model (base model and the classifier or decoder) on the new finetuning data (again by first applying language-specific sparse updates before forward pass as described in \S \ref{subsec:background}). Sparse finetuning ensures that the updated model is minimally different from the deployed model (roughly 5-10\% parameters are kept trainable).

During inference, we apply the sparse update matrix of the test language before the forward pass.

\vspace{0.5ex}
\noindent
\emph{\textbf{Language-specific Adapter Finetuning (LAFT)}}
Here the goal is to split the model into language-specific (adapters) and language-agnostic (base model) parts. For the inception stage, we first take the deployed model from FFT and insert (randomly initialized) adapters in each layer of the network. We train the adapter layers and the classifier or decoder on inception stage data for all languages for the task and then create $N_L$ copies of the trained adapters (one for each language). The $i^{th}$ copy is again finetuned (with the base model frozen) on inception stage data but this time only using the data for the $l_{i}$ language. This gives $N_L$ language specific adapters, a shared base model and a shared classifier or decoder (\textit{inception stage} diagram at appendix \ref{sec:appendix_laft_full}). During inference, one can simply swap to the adapter corresponding to the test language.

For the continuation stage, given access to new finetuning data in $l_j$ language, we use the $j^{th}$ adapter during the forward pass and update it using gradient descent. Usually in adapter-based strategies, the shared base model is kept frozen. However, in our setting, we would want to encourage knowledge transfer between languages. At the same time, we would want to make minimum changes to the shared base model to avoid losses due to forgetting. We balance the two goals by keeping the base model trainable but with a much lower learning rate (compared to the adapter layers). Since in the inception stage, the model has associated language-specific learning with the adapters and language-agnostic learning with the base model, this would incentivize the model to not overfit the shared base model on the language regularities of the new data. We find that LAFT shows improved behaviour compared to FFT \& SFT.
\begin{figure}[t!]
\centering
\includegraphics[width=7.5cm]{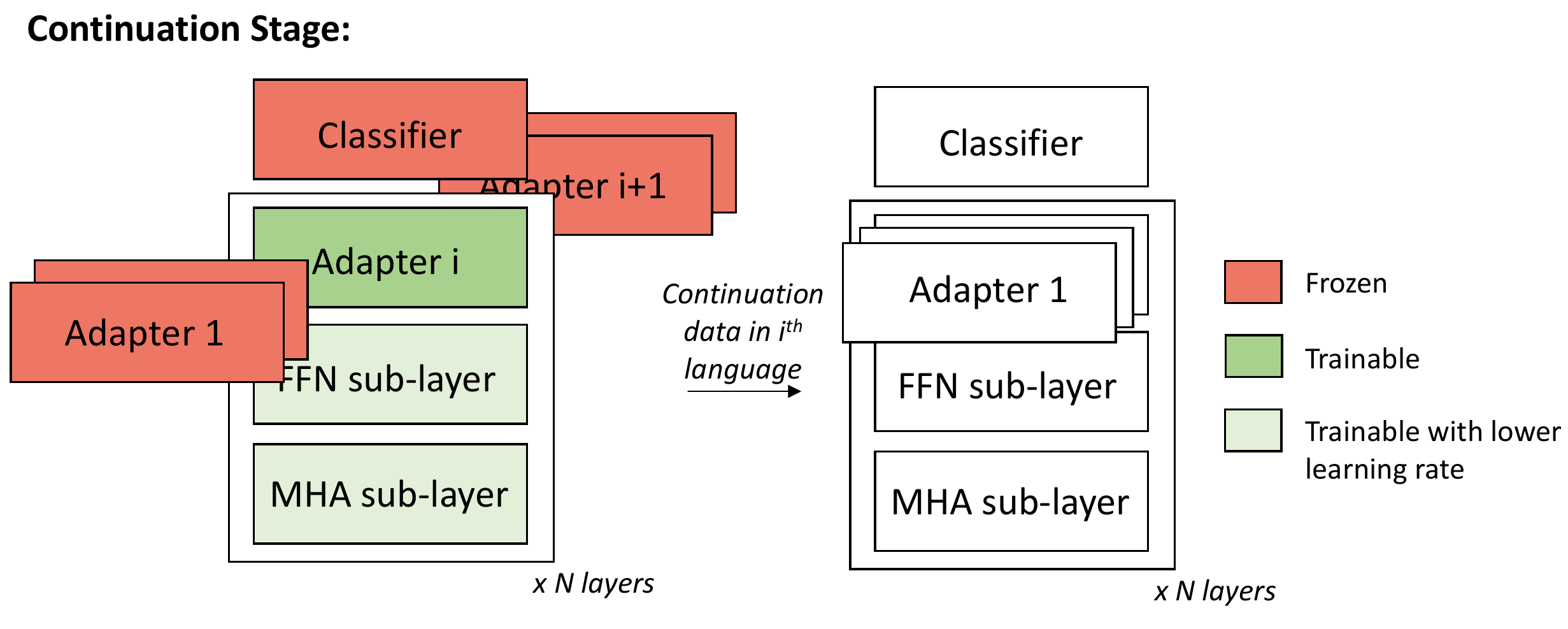}
\caption{Continued finetuning using our proposed method, LAFT-URIEL. Here we split the model into language-agnostic (base model) and language-specific parts (adapters). The language-agnostic part is trained with a lower learning rate compared to the language-specific part of the network. Lowering of the learning rate is dynamically decided based on composition of the continuation stage data. This helps in sharing performance gains across languages while reducing model's tendency to become overspecialized on the language of the new data. \S \ref{subsec:improved_pipelines} for more details.}
\label{fig:laft}
\end{figure}

\vspace{0.5ex}
\noindent
\emph{\textbf{LAFT using URIEL distances (LAFT-URIEL)}}
\begin{figure*}[t!]
\centering
\includegraphics[width=14cm]{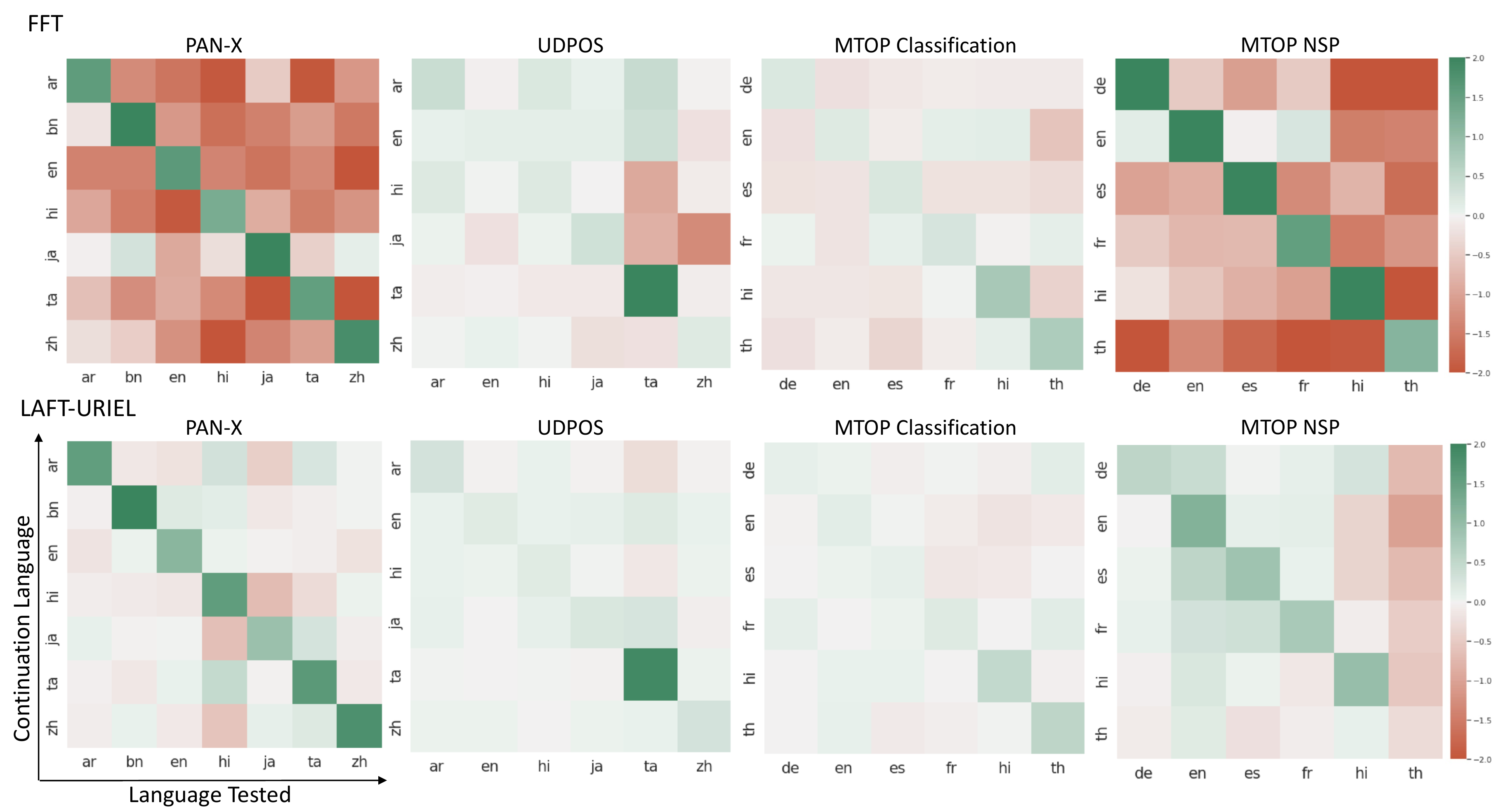}
\caption{Performance change heatmaps (see \S \ref{sec:result_fft_laft}) using  FFT (top) and the LAFT-URIEL (bottom) strategies on all four tasks, plotted on the same scale on the right end (+2 to -2\%). Here each red cell indicates that there was a loss in performance on the language denoted by the column, after continued finetuning on the language denoted by the row. The colour intensity corresponds to the magnitude of change. Our proposed method, LAFT-URIEL, greatly improves upon the baseline (FFT) by reducing both the number and intensity of red cells in the heatmap.}
\label{fig:comparison_hm}
\end{figure*}
We argue that the selection of learning rate (LR) of the base model should be made based on the language-wise composition of the new finetuning data. If we know that the new data is skewed towards a language which is very ``different'' from the remaining languages, then keeping the LR low would be the desired choice as it is very unlikely that finetuning on this new data would lead to shared gains in performance.

We use URIEL syntactic distance as a measure of similarity between different languages. We calculate the LR of the base model by dividing the LR of the adapter layers (kept the same across languages) by a division factor. For continuation stage with data in $l_i$ language, the division factor is computed as a linear function of the average syntactic distance of $l_i$ from $\{l_1, l_2, ..., l_{N_L}\}\setminus{\{l_i\}}$. We show this calculation for the MTOP NSP task in Table \ref{table:laft_uriel}. We call this strategy LAFT-URIEL and its continuation stage diagram is represented in Fig \ref{fig:laft}.

\begin{table}[t!]
\centering
\small{\begin{tabular}{@{}ccc@{}}
\toprule
  \textbf{Lang} & \textbf{Avg syn. distance} & \textbf{LR of base model (in $10^{-5}$)} \\ 
 \hline
 en & 0.405 &  $5/35$ \\ 
 de & 0.435 &  $5/50$\\
 es & 0.415 &  $5/40$\\
 fr & 0.422 &  $5/45$\\ 
 hi & 0.498 &  $5/80$\\
 th & 0.540 & $5/100$\\\bottomrule
\end{tabular}}
\caption{Learning rate (LR) of the base model changing with the language of the continuation stage data for MTOP NSP using LAFT-URIEL strategy (\S \ref{subsec:improved_pipelines}).}
\label{table:laft_uriel}
\end{table}
\section{Comparison of Finetuning Strategies}
In this section, we use the metrics defined in \S \ref{sec:baseline} to compare the four finetuning strategies (FFT, SFT, LAFT, LAFT-URIEL) on the four tasks described in \S \ref{sec:problem_setup}. It is important to note that the absolute language-wise performances of the deployed models in SFT, LAFT and LAFT-URIEL cases are at par or slightly greater than the same for FFT (see appendix \ref{sec:appendix_performance}). Since our metrics are computed over changes in performance, the above fact ensures that the comparison is fair (or slightly favourable towards FFT). 
\\
We ask the following research questions:\footnote{results in main paper are using mBERT, see appendix \ref{sec:appendix_xlm} for results using XLM-RoBERTa}\\
\textit{\textbf{1.} How does the behaviour of our proposed strategy differ from that of the na\"ive baseline} (\S \ref{sec:result_fft_laft}),\\
\textit{\textbf{2.} Do parameter-efficient finetuning methods improve spread of gains after while constraining the losses in our setup?} (\S \ref{sec:result_numlangs} and \S \ref{sec:result_loss}), \\
\textit{\textbf{3.} How does our proposed strategy perform when there are multiple continuation stages?} (\S \ref{sec:result_mult})
\subsection{Behaviour of FFT and LAFT-URIEL}
\label{sec:result_fft_laft}
We construct heatmaps to visualize the performance changes observed after a continuation stage (Figure \ref{fig:comparison_hm}). Here each row corresponds to a continuation stage between the $\mathrm{t=0}$ and the $\mathrm{t=1}$ models where the new finetuning data is only present in the language corresponding to the row index. In other words, given the same deployed model, each row corresponds to a different updated model. The column index corresponds to the language used to evaluate the updated model.
We present heatmaps for both the FFT (top row) and the LAFT-URIEL (bottom row) strategies in Figure \ref{fig:comparison_hm}. For the FFT baseline, the diagonal entries are highly positive while many of the off-diagonal entries are negative across all four tasks. This indicates that the model is overfitting on the language specificities of the new finetuning data leading to degraded generalization capabilities across the remaining languages.

For LAFT-URIEL, we observe improved behaviour across all four tasks. The green cells in the LAFT-URIEL are much more evenly spread and higher in number compared to FFT. We also notice that the intensity of the red cells have reduced significantly. This behaviour is much closer to the ideal behaviour than FFT\footnote{similar improvements using XLM-RoBERTa, appendix \ref{sec:appendix_xlm}}. In the subsequent subsections, we will quantify these observations using the metrics proposed in section \ref{sec:baseline}.

\subsection{Measuring Spread of Gains}
\label{sec:result_numlangs}
\begin{figure}[t!]
\centering
\includegraphics[width=7.5cm]{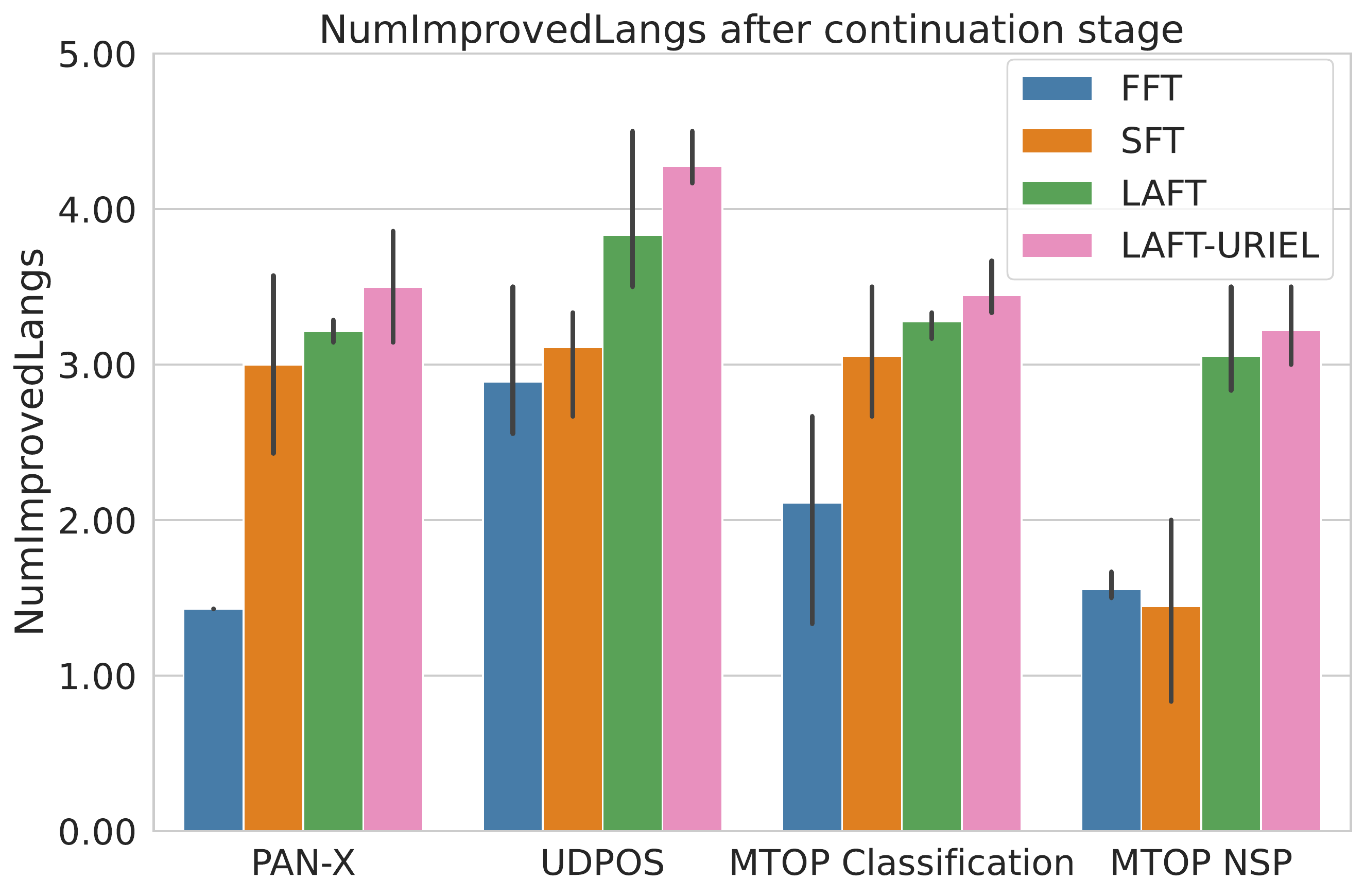}
\caption{$\mathrm{NumImprovedLangs}$ across tasks (std-dev in black). Higher the better. LAFT and LAFT-URIEL, improve spread of gains compared to both FFT and SFT. See \S \ref{sec:result_numlangs} for more details.}
\label{fig:avg_langs}
\end{figure}
We plot $\mathrm{NumImprovedLangs}$ on all four tasks in Figure \ref{fig:avg_langs}. We see significant gains over the na\"ive FFT baseline using SFT across all tasks but UDPOS, indicating that SFT is indeed a stronger baseline for our setup. Both LAFT and LAFT-URIEL improve upon the SFT strategy on this metric. The gains of LAFT-URIEL over LAFT indicates that using URIEL syntactic distance to dynamically compute the learning rate of the base model given the composition of the continuation stage data helps in improving positive transfer across languages. With FFT, an average of only \textbf{32.48\%} of task languages could improve after the $\mathrm{t=0}$ to $\mathrm{t=1}$ continuation stage. This number increases to \textbf{58.08\%} using LAFT-URIEL, therefore suggesting that \emph{majority of languages are expected to improve using our proposed strategy when a multilingual model is further finetuned on new language-specific data for the same task}. 

\subsection{Comparing Losses Incurred}
\label{sec:result_loss}
\begin{figure}[t!]
\centering
\includegraphics[width=7.5cm]{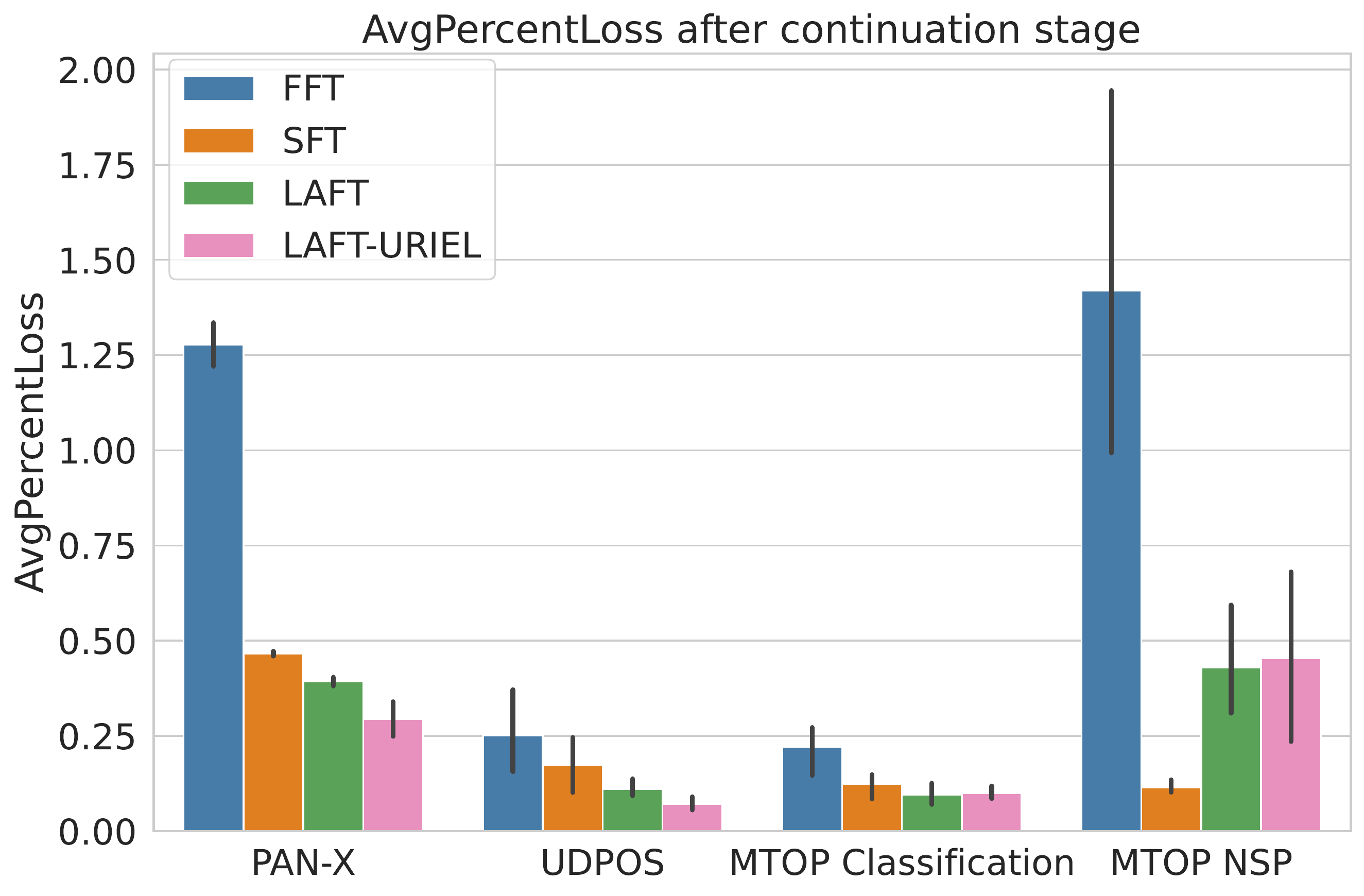}
\caption{$\mathrm{AvgPercentLoss}$ across four tasks (std-dev in black). Lower the better. LAFT-URIEL shows significant improvement over FFT and SFT on 3 tasks. SFT doesn't meet the constraints (Table \ref{table:all_metrics}) for MTOP NSP. See \S \ref{sec:result_loss} for more details.}
\label{fig:avg_loss}
\end{figure}
\begin{table*}[h]
\centering
\scalebox{0.85}{\begin{tabular}{@{}ccccccccc@{}}
\toprule
  \multirow{2}{*}{\textbf{Strategy}} & \multicolumn{2}{c}{\textbf{PAN-X}} 
  & \multicolumn{2}{c}{\textbf{UDPOS}} & \multicolumn{2}{c}{\textbf{MTOP Classification}} & \multicolumn{2}{c}{\textbf{MTOP NSP}} \\ 
  \cline{2-9}
 & \small{$\mathbf{MaxRatio}$} & \small{$\mathbf{SumRatio}$} & \small{$\mathbf{MaxRatio}$} & \small{$\mathbf{SumRatio}$}& \small{$\mathbf{MaxRatio}$} & \small{$\mathbf{SumRatio}$}& \small{$\mathbf{MaxRatio}$} & \small{$\mathbf{SumRatio}$}\\
 \hline
  FFT & 1.23 \tiny{$\pm$ 0.25} & \unhighlight{0.50\tiny{$\pm$ 0.11}}  & 6.85\tiny{$\pm$ 2.26} & 5.10\tiny{$\pm$ 3.23} & 1.19\tiny{$\pm$ 1.02} & \unhighlight{0.82\tiny{$\pm$ 0.8}} & 1.16\tiny{$\pm$ 0.48} & \unhighlight{0.48\tiny{$\pm$ 0.27}} \\ 
  SFT & 1.26\tiny{$\pm$ 0.19} & \unhighlight{0.76\tiny{$\pm$ 0.58}} & 4.95\tiny{$\pm$ 2.31} & 2.58\tiny{$\pm$ 0.95} & 2.73\tiny{$\pm$ 0.72} & 2.62\tiny{$\pm$ 1.12} & \unhighlight{0.54\tiny{$\pm$ 0.13}} & \unhighlight{0.36\tiny{$\pm$ 0.16}} \\
  LAFT & 4.62\tiny{$\pm$ 0.53} & 2.55\tiny{$\pm$ 0.51} & 11.72\tiny{$\pm$ 3.17} & 13.25\tiny{$\pm$ 4.23} & 9.65\tiny{$\pm$ 1.45} & 17.64\tiny{$\pm$ 3.56} & \highlight{1.26\tiny{$\pm$ 0.31}} & \highlight{1.70\tiny{$\pm$ 0.26}} \\
  LAFT-URIEL & \highlight{6.30\tiny{$\pm$ 0.68}} & \highlight{4.10\tiny{$\pm$ 0.76}} & \highlight{$\infty$} & \highlight{$\infty$} & \highlight{12.04\tiny{$\pm$ 3.31}} & \highlight{20.82\tiny{$\pm$ 4.54}} & 1.18\tiny{$\pm$ 0.22} & 1.48\tiny{$\pm$ 0.34} \\ \bottomrule
\end{tabular}}
\caption{$\mathrm{MaxRatio}$ and $\mathrm{SumRatio}$ (\S \ref{sec:baseline}) for all four strategies across all four tasks. Higher the better.  Values  highlighted  in  red  are unfavorable and indicates that the updated model is worse than the previous version on an overall aggregated metric. LAFT-URIEL shows close to ideal behaviour on all tasks.}
\label{table:all_metrics}
\end{table*}
\begin{table*}[h]
\centering
\scalebox{0.85}{\begin{tabular}{@{}ccccccccc@{}}
\toprule
  \multirow{2}{*}{\textbf{Strategy}} & \multicolumn{2}{c}{\textbf{PAN-X}} 
  & \multicolumn{2}{c}{\textbf{UDPOS}} & \multicolumn{2}{c}{\textbf{MTOP Classification}} & \multicolumn{2}{c}{\textbf{MTOP NSP}} \\ 
  \cline{2-9}
 & \small{$\mathbf{\%Loss}$} & \small{$\mathbf{+veLangs}$} & \small{$\mathbf{\%Loss}$} & \small{$\mathbf{+veLangs}$}& \small{$\mathbf{\%Loss}$} & \small{$\mathbf{+veLangs}$}& \small{$\mathbf{\%Loss}$} & \small{$\mathbf{+veLangs}$}\\
 \hline
  FFT \small{(\textsc{l2h})} & 1.737 & 1 & 0.839 & 1 & 0.378 & 3 & 5.459 & \highlight{5} \\ 
  LAFT-URIEL \small{(\textsc{l2h})} & \highlight{0.505} & \highlight{4} & \highlight{0.182} & \highlight{2} & \highlight{0.210} & \highlight{4} & \highlight{0.778} & 2 \\ \midrule
  FFT \small{(\textsc{h2l})} & 1.406 & 1 & 0.598 & 1 & 0.283 & 3 & 2.334 & \highlight{2} \\ 
  LAFT-URIEL \small{(\textsc{h2l})} & \highlight{0.613} & \highlight{4} & \highlight{0.123} & \highlight{4} & \highlight{0.192} & \highlight{4} & \highlight{0.773} & \highlight{2} \\
  \bottomrule
\end{tabular}}
\caption{Magnitude of $\mathrm{AvgPercentLoss}$ (denoted by $\mathrm{\%Loss}$) and $\mathrm{NumImprovedLangs}$ (denoted by $\mathrm{+veLangs}$) calculated on the \emph{worst-case continuation stage} for FFT and LAFT-URIEL after multiple continuation stages. \textsc{h2l} and \textsc{l2h} represent two trajectories we consider (\S \ref{sec:result_mult}). LAFT-URIEL ensures that the losses are constrained even after multiple continuation stages.}
\label{table:mult}
\end{table*}
We plot $\mathrm{AvgPercentLoss}$ for all the four strategies on all four tasks in Figure \ref{fig:avg_loss}. We again observe considerable improvement over the FFT baseline using SFT. LAFT and LAFT-URIEL improves upon SFT across all tasks except MTOP NSP. Upon closer analysis, we find out that both the magnitude of gains and losses for SFT in this task are severely constrained, because of which the changes in language-wise performance are close to zero. This is also reflected in Table \ref{table:all_metrics}, where the $\mathrm{MaxRatio}$ and $\mathrm{SumRatio}$ values for SFT are non-ideal for MTOP NSP. We believe that this might be due to the fact that the language-specific update matrices are only available for the encoder of the network because of which the task-language decomposition is hampered. LAFT and LAFT-URIEL satisfy the minimum criteria (value $\geq 1$) for all tasks (Table \ref{table:all_metrics}). For LAFT-URIEL in the UDPOS task, there are continuation stages where none of the languages incur a loss in performance because of which the average of the two ratios come out to be $\infty$. This is a significant improvement in behaviour compared to both SFT and FFT. \emph{LAFT-URIEL reduces the magnitude of losses incurred by around \textbf{78\%} relative on an average compared to FFT.}
\subsection{Multiple Continuation Stages}
\label{sec:result_mult}
We also evaluate LAFT-URIEL on multiple continuation stages, sequentially performed one after another. We consider two trajectories for sequential finetuning: high-resource to low-resource (\textsc{h2l}) and low-resource to high-resource (\textsc{l2h}) languages (inspired by \citet{m2022cross}) based on number of examples in the training data for a given task\footnote{We believe that these ordered trajectories would be more challenging compared to a random trajectory since it is easier for the model to overfit/forget the low resource languages} and report the metrics on the \emph{worst-case continuation stage} in a trajectory. Given a trajectory of the form $\xrightarrow{\{l_{i_1}\}}\xrightarrow{\{l_{i_2}\}}...\xrightarrow{\{l_{i_{N_L}}\}}$, we define the \emph{worst-case continuation stage}, $\mathrm{t_w}$, as follows: 
\begin{gather}
    \mathrm{model(t=t_w-1)} \xrightarrow{\{l_{t_w}\}} \mathrm{model(t=t_w)} \\
    where\ \mathrm{t_w} = \underset{\mathrm{t}}{\mathrm{argmax}}\ \mathrm{AvgPercentLoss(t)}
\end{gather}
i.e the continuation stage where the $\mathrm{AvgPercentLoss}$ was maximum in the trajectory.
We report the metrics on the worst-case continuation stage in the Table \ref{table:mult}.
We observe that our proposed strategy consistently reports a value $\leq1\%$ for worst-case $\mathrm{AvgPercentLoss}$ across all tasks. Also, there are $>1$ languages with improved performance, even after the worst-case continuation stage. This is a strong result which indicates that our finetuning strategy is able to control losses and spread gains even after multiple continuation stages.
\\ \\
We refer the readers to the Appendix for further experiments which aim to understand (1) how the size of adapter layers affect the performance of the LAFT strategy (\S \ref{sec:appendix_ablation}), (2) the effect of continued finetuning on closely related languages (\S \ref{sec:appendix_close}), and (3) the variance in cross-lingual transfer across seeds, tasks and encoders (\S \ref{sec:appendix_challengs})

\section{Related Works}
\textbf{Continual Learning:}
A large body of work in the continual learning literature is focused on the \textit{task incremental} setting \citep{de2021continual} where the goal is to sequentially introduce new tasks to the network. Elastic weight consolidation \citep{kirkpatrick2017overcoming} is one of the most widely used algorithms for this setting, however, it assumes that the old training data is available for computing the regularization term. \citet{chen-etal-2020-recall} proposes the RecAdam optimizer which further approximates the computation of the Fisher information matrix so that there is no need for having access to the old training data. The resulting optimizer imposes a quadratic penalty on the difference of the current values and the old values of the parameters of the network. A similar penalty is also already incorporated in the SFT strategy ($L_1$ norm of the difference in this case). 
Recent works in studying multilingual modelling from a continual learning perspective include works of \citet{m2022cross, yang2022towards} which study incrementally adding task data in \textit{unseen languages} and \citet{berard2021continual, garcia2021towards} on extending the language capacity of MT models; both very different from our setup.
\\ \\
\textbf{Parameter-efficient finetuning:}
Parameter-efficient finetuning methods such as adapters have shown promise in multi-task continual learning setups \citep{ke-etal-2021-adapting} as well as zero-shot cross-lingual transfer \citep{pfeiffer-etal-2020-mad, ansell-etal-2021-mad-g}. Recent works \citep{ponti2022combining} utilize such methods to decompose task learning into underlying skill learning and allocation.

\section{Conclusion}
In this paper we introduce and study the problem of \textbf{Continual Multilingual Learning (CML)} where a multilingual model is continually updated using new data from a subset of the languages at a time. We observe that unconstrained updates to the model can lead to drastic losses for a subset of the languages, especially those not covered during an update. We propose  LAFT-URIEL, a parameter-efficient finetuning stragegy which uses linguistic information to effectively balance overfitting and knowledge sharing across different languages, resulting in 25\% increase in the proportion of task languages whose performances improve during an update while achieving 78\% relative decrease in average magnitude of losses on the remaining languages.


\section{Limitations and Future Work}
Since this is one of the first studies on understanding the effects of continued finetuning of multilingual models, the focus of this paper was to lay the groundwork by establishing the experimental setting on a set of representative NLP tasks and languages. The resulting set of languages chosen in our setup for evaluation (en, hi, bn, zh, ta, ja, ar, de, es, fr, th), although diverse, are still relatively higher resource. Extending the analysis to languages which were severely underrepresented (or even absent) during the pretraining of the underlying model may provide interesting insights and would be an important future work to pursue.

\section*{Acknowledgements}

The authors would like to express their gratitude to Sebastian Ruder, Srini Narayanan and Rachit Bansal for helpful overall feedback, Bidisha Samanta for discussions on problem formulation, Jon Clark for feedback on methodologies and Nitish Gupta for insightful comments on the metrics.
\bibliography{custom}
\bibliographystyle{acl_natbib}

\appendix

\onecolumn
\section{Experimental Settings}
\label{sec:appendix_settings}
All of our experiments are performed on four NVIDIA$^{\tiny{\textregistered}}$ A100-SXM4-40GB GPUs. Our implementation uses PyTorch \citep{NEURIPS2019_9015}, the Transformers library \citep{wolf2019huggingface}, AdapterHub \citep{pfeiffer2020adapterhub} and Composable-SFT \cite{ansell-etal-2022-composable}. We use \href{https://huggingface.co/bert-base-cased}{bert-base-cased} and \href{https://huggingface.co/xlm-roberta-base}{xlm-roberta-base} checkpoints to initialize our models. For the seq2seq task, both the encoder and decoder are initialized by the above multilingual checkpoints, as suggested by \citet{rothe2020leveraging}. The new cross-attention terms in the decoder are initialized from scratch.

We present the hyperparameters selected for each finetuning strategy in Table \ref{table:hyperparams}. We use the AdamW optimizer \citep{loshchilov2017decoupled,kingma2014adam} with weight decay of 1e-5 for each pipeline and perform a search across 3 learning rate values (2, 5 and 8 $\times10^{-5}$) for each strategy and finetuning stage and select the best performing model using the dev set. For SFT continuation, we experiment with different percentages of the number of trainable parameters in the network and report the best configuration. We also find that freezing the layer norm parameters while sparsely finetuning the entire model (both base model and classifier/decoder) for SFT leads to improvement in behaviour for our task.

We use a batch size of 64 for PAN-X and UPDOS, 128 for MTOP Classification and 96 for MTOP semantic parsing.
\begin{table*}[h]
\centering
\small{\begin{tabular}{@{}cSLLLL@{}}
\toprule
  \multicolumn{2}{c}{\textbf{Strategy}} & \textbf{PAN-X} 
  & \textbf{UDPOS} & \textbf{MTOP Class.} & \textbf{MTOP NSP} \\ 
 \hline
  \multirow{2}{*}{FFT} & \textit{Inception} & lr: 2e-5, num epochs: 10 & lr: 2e-5, num epochs: 10 & lr: 2e-5, num epochs: 10 & lr: 8e-5, num epochs: 40  \\\cline{2-6}
  & \textit{Continuation} & lr: 2e-5, num epochs: 10 & lr: 2e-5, num epochs: 10 & lr: 2e-5, num epochs: 10 & lr: 2e-5, num epochs: 20  \\ \midrule\midrule
  \multirow{2}{*}{SFT} & \textit{Inception} & lr: 2e-5, ft epochs: 3, st epochs: 10, base trainable params: 14155776 & lr: 2e-5, ft epochs: 3, st epochs: 10, base trainable params: 14155776 & lr: 2e-5, ft epochs: 3, st epochs: 10, base trainable params: 14155776 & lr: 8e-5, ft epochs: 10, st epochs: 40, base trainable params: 88926720  \\\cline{2-6} 
  & \textit{Continuation} & lr: 2e-5, ft epochs: 3, st epochs: 10, base trainable params: 7077888 & lr: 2e-5, ft epochs: 3, st epochs: 10, trainable params: 7077888 & lr: 2e-5, ft epochs: 3, st epochs: 10, base trainable params: 7077888 & lr: 2e-5, ft epochs: 10, st epochs: 20, base trainable params: 14155776  \\ \midrule\midrule
  \multirow{3}{*}{LAFT} & \textit{Inception (shared adapters)} & lr: 2e-5, num epochs: 10 & lr: 2e-5, num epochs: 10 & lr: 2e-5, num epochs: 10 & lr: 8e-5, num epochs: 40  \\ \cline{2-6}
  & \textit{Inception (lang-specific adapters)} & lr: 2e-5, num epochs: 10 & lr: 2e-5, num epochs: 10 & lr: 2e-5, num epochs: 10 & lr: 2e-5, num epochs: 20  \\ \cline{2-6}
  & \textit{Continuation} & lr: 2e-5, num epochs: 10, div factor: 10 & lr: 2e-5, num epochs: 10, div factor: 10& lr: 2e-5, num epochs: 10, div factor: 10 & lr: 5e-5, num epochs: 20, div factor: 50  \\ \midrule
  \bottomrule
\end{tabular}}
\caption{Best hyperparameters for each strategy, stage and task. LR denotes learning rate of the entire model for FFT/SFT and for the adapter layers in LAFT. Div factor denotes the division factor used to calculate the learning rate of the base model relative to those of adapter layers for the LAFT strategies. For SFT, FT epochs denote the number of pilot training epochs used to select the top-$k$ parameters which will be kept trainable in the subsequent sparse finetuning epochs (denoted by ST).}
\label{table:hyperparams}
\end{table*}

\section{Dataset Details}
\label{sec:appendix_dataset}
We provide the language-codes to language mapping in Table \ref{table:lang_codes}.
\begin{table}[h]
\centering
\small{\begin{tabular}{@{}lc@{}}
\toprule
  \textbf{ISO 639-1 Language Code} & \textbf{Language} \\ 
 \hline
 en & English \\
 hi & Hindi \\
 bn & Bengali \\
 zh & Chinese \\
 ta & Tamil \\
 ja & Japanese \\
 ar & Arabic \\
 de & German \\
 es & Spanish \\
 fr & French \\
 th & Thai \\\bottomrule
\end{tabular}}
\caption{Language-code mapping for all the languages covered in our study.}
\label{table:lang_codes}
\end{table}
We also present the training data distribution across different languages and the evaluation metric used for the four tasks we consider in our study in Table \ref{table:dataset_statistics}.
\begin{table}[h]
\centering
\small{\begin{tabular}{@{}llll@{}}
\toprule
  \textbf{Dataset} & \textbf{Evaluation Metric} & \textbf{Ratio of Training Examples}  & \textbf{Total} \\ 
 \hline
 PAN-X & Macro-F1 & 1 (en) : 0.25 (hi) : 0.5 (bn) : 1 (zh) : 0.75 (ta) : 1 (ja) : 1 (ar) & 110k \\ 
 UDPOS &  Macro-F1 & 1 (en) : 0.62 (hi) : 0.33 (ja) : 0.019 (ta) : 0.89 (zh) : 0.28 (ar) & 67k \\
 MTOP Class. & Macro-F1 & 1 (en) : 0.85 (de) : 0.69 (es) : 0.75 (fr) : 0.72 (hi) : 0.68 (th) & 74k \\
 MTOP NSP &  Exact Match Acc. (EMA) & 1 (en) : 0.85 (de) : 0.69 (es) : 0.75 (fr) : 0.72 (hi) : 0.68 (th) & 74k\\ \bottomrule
\end{tabular}}
\caption{Languages covered for the tasks we consider in our setup. }
\label{table:dataset_statistics}
\end{table}

To evaluate performance on token level tasks (PAN-X and UPDOS), we use seqeval toolkit \citep{seqeval}. We obtain these two datasets from the XTREME benchmark \citep{hu2020xtreme}.

For UDPOS, we consider the following POS tags: \texttt{['ADJ', 'ADP', 'ADV', 'AUX', 'CCONJ', 'DET', 'INTJ', 'NOUN', 'NUM', 'PART', 'PRON', 'PROPN', 'PUNCT', 'SCONJ', 'SYM', 'VERB', 'X' ]}\\
For PAN-X we consider the following NER tags: \texttt{['O', 'B-PER', 'I-PER', 'B-ORG', 'I-ORG', 'B-LOC', 'I-LOC']}\\
MTOP Domain classification is a 11-way sentence classification task. The dataset has 117 intents and 78 slots for the semantic parsing task.

\section{Diagrammatic View of the LAFT Strategy}
\label{sec:appendix_laft_full}
\begin{figure*}[h!]
\centering
\includegraphics[width=\textwidth]{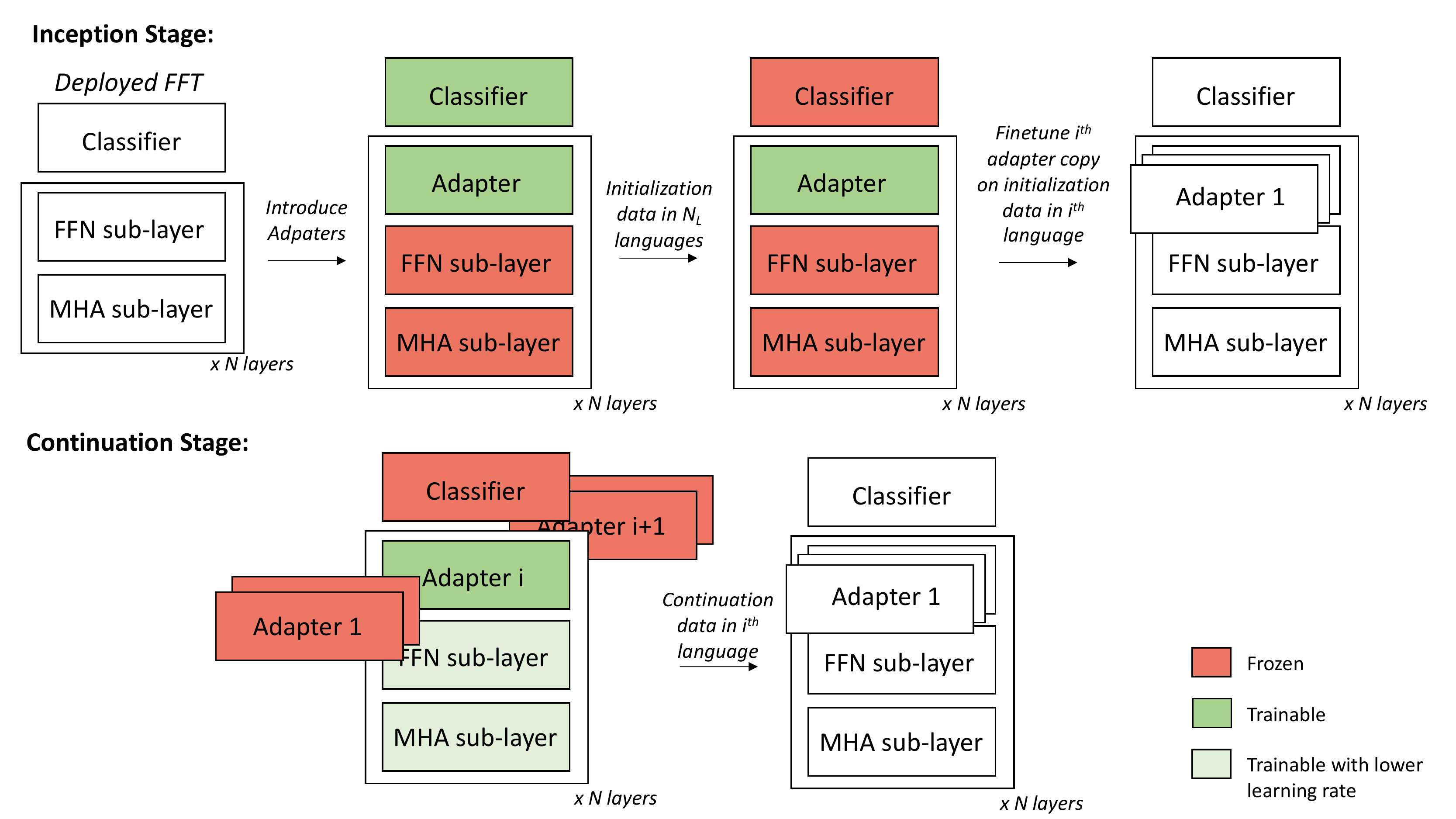}
\caption{Full diagrammatic representation of the LAFT strategy}
\label{fig:laft_full_appendix}
\end{figure*}

\section{Model Performance Comparison after Inception Stage}
\label{sec:appendix_performance}
\begin{table}[h]
\centering
\small{\begin{tabular}{@{}llll@{}}
\toprule
  \textbf{Dataset} & \textbf{FFT} & \textbf{SFT} & \textbf{LAFT} \\ 
 \hline
 PAN-X & 81.00 & \highlight{82.50} & 81.65 \\ 
 UDPOS & 88.67 & 88.97 & \highlight{89.15} \\
 MTOP Classification & 96.71 &  97.06 & \highlight{97.08} \\
 MTOP NSP & 59.41 &  60.61 & \highlight{61.16} \\\bottomrule
\end{tabular}}
\caption{Macro average performance (avg taken across languages) after the inception stage for each strategy. LAFT and LAFT-URIEL share the same model after inception. Training data for the inception stage is 50\% training data for the task in each language.}
\label{table:inception_performance}
\end{table}
In Table \ref{table:inception_performance}, we report the model performance after inception stage for each strategy (macro average across languages). Variance across different strategies is low. LAFT most often produces the best performing model after inception. Since our metrics are defined on changes in performance, the above fact ensures that our analysis is fair (or slightly favourable) towards the baseline since the first deployed model for the LAFT strategy has less scope of improvement after continuation.

\section{Heatmaps for XLM-RoBERTa}
\label{sec:appendix_xlm}

\begin{figure*}[h!]
\centering
\includegraphics[width=\textwidth]{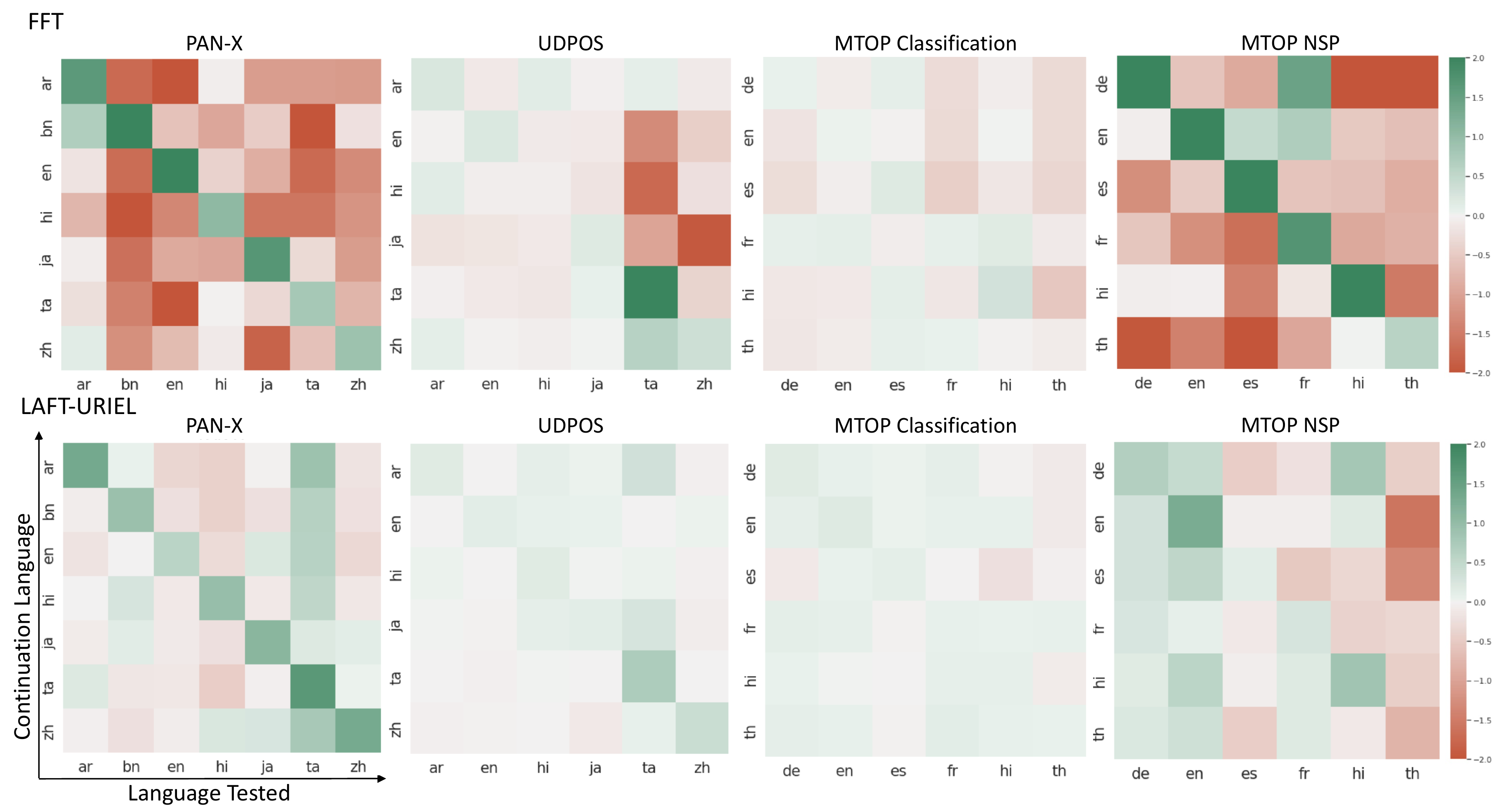}
\caption{Performance change heatmaps for the FFT (top) and the LAFT-URIEL (bottom) using XLM-RoBERTa initialization on all four tasks. LAFT-URIEL again shows improved behaviour with more green cells and reduced intensity of red cells.}
\label{fig:comparison_hm_xlm}
\end{figure*}

We compare performance change heatmaps for FFT and LAFT-URIEL across all four tasks in Figure \ref{fig:comparison_hm_xlm}. We notice the same improved behaviour as observed in \S \ref{sec:result_fft_laft} using the mBERT initialization. This suggests that gains observed using our proposed strategy are consistent across different model initializations. 
\section{Size of Adapters for the LAFT Strategy}
\label{sec:appendix_ablation}
We perform experiments to study how the size of the adapter layers in the LAFT strategy affects the model's ability to (1) share the gains due to continued finetuning across languages and (2) control for language specific losses. For this, we calculate $\mathrm{NumImprovedLangs}$ and $\mathrm{AvgPercentLoss}$ on the PAN-X dev set for the $\mathrm{t=0}$ to $\mathrm{t=1}$ transition, by varying the size of the adapter layers. The size of an adapter layer is controlled by the bottleneck dimension, $b_{dim}$ (throughout our experiments, we use $b_{dim}=48$ for both LAFT and LAFT-URIEL). We present results in table \ref{table:bdim}.
\begin{table}[h!]
\centering
\small{\begin{tabular}{@{}ccc@{}}
\toprule
  \textbf{$b_{dim}$} & $\mathbf{AvgPercentLoss}$ & $\mathbf{NumImprovedLangs}$ \\ 
 \hline
 48 & \highlight{0.40} &  \highlight{3.14} \\ 
 24 & 0.46 &  \highlight{3.14} \\
 12 & 0.41 &  2.85 \\\bottomrule
\end{tabular}}
\caption{$\mathrm{AvgPercentLoss}$ and $\mathrm{NumImprovedLangs}$ for the LAFT strategy by varying $b_{dim}$ on PAN-X dev set}
\label{table:bdim}
\end{table}

As we reduce $b_{dim}$ for LAFT, we see there is either a hit in positive transfer or increase in magnitude of loss. Since the capacity of the language-specific part of the network is decreased, the model might be relying more on the language-agnostic part during continued finetuning, making it more susceptible to overfit and degrade its multilingual capabilities. We therefore use $b_{dim}=48$ which correspond to roughly $3.5\%$ of the base model parameters for the tagging task. In comparison, the language-specific update matrices used in the SFT strategy corresponds to roughly 4\% of the base model parameters, and results in an $\mathrm{AvgPercentLoss}$ of 0.47 and $\mathrm{NumImprovedLangs}$ of 2.85 on the same dev set used in this analysis.

\section{Effect of Continued Finetuning on Closely-related Languages}
\label{sec:appendix_close}
In our setting, one may assume that continued finetuning on new task-specific data in some language should also benefit the languages which are most closely related to it. To study this, we use the URIEL syntactic distance metric to find the two languages closest to a given language $L$, denoted as $L_1$ (closest) and $L_2$ (second-closest). We expect that continued finetuning in language $L$ would benefit $L_1$ more than $L_2$ since $L_1$ is more closely related to $L$.

To test this, we calculate the percentage of times change in performance in $L_1$ is greater than that of $L_2$ by varying $L$ for a given task. We report the numbers in table \ref{table:mmhmm} for LAFT-URIEL and find that they support our hypothesis that the language closest to $L$ is more likely to experience more favorable performance change than the second closest language.

\begin{table}[h!]
\centering
\small{\begin{tabular}{@{}cc@{}}
\toprule
  \textbf{Task} & \textbf{\% of times performance change in $L_1$ $\geq$ change in $L_2$}\\ 
 \hline
 PAN-X & 85.70  \\ 
 UDPOS & 66.66  \\
 MTOP Classification & 83.33 \\
 MTOP NSP & 83.33 \\ \bottomrule
\end{tabular}}
\caption{Percentage of times performance change of closest language to $L$ is greater than second closest language to $L$ after continued finetuning in $L$ for different tasks using LAFT-URIEL}
\label{table:mmhmm}
\end{table}

\section{Quantifying Variance in Language-wise Performance Change after Continued Finetuning}
\label{sec:appendix_challengs}
Throughout different experiments in our setting, we observe significant variation on the order of languages which are most improved to most degraeded after continued finetuning on new data for a given language. We attempt to quantify this by analyzing how the behaviour of the model changes when we vary (1) random seed keeping task constant (2) encoder keeping task constant (3) task keeping dataset constant. 

To do this, we first construct performance change heatmaps after performing the above-listed variations and then find out the order in which language-wise performance is negatively impacted for a given continuation stage (i.e sorting \% change in performance across languages). We compare this order with the original order by computing the edit distance between the two. High edit distance would indicate that order of performance change is very sensitive to the factors we are changing in this analysis. We present the following results (evaluated using FFT):

\begin{itemize}
    \item \textbf{Varying random seed for same task}: Average edit distance after changing seed for the four tasks is as follows: PANX: 4 ($N_L$=7); UDPOS: 3.66 ($N_L$=6); MTOP Classification: 4.16 ($N_L$=6); MTOP NSP: 3.55 ($N_L$=6)
    \item \textbf{Varying task for same dataset (MTOP)}: Average edit distance between  rows of heatmaps obtained after MTOP Classification and MTOP NSP is 3.66 ($N_L$=6)
    \item \textbf{Varying encoder (mBERT or XLM-RoBERTa) for same task}: Average edit distance after changing encoder for the four tasks is as follows: PANX: 4.57 ($N_L$=7); UDPOS: 3.33 ($N_L$=6); MTOP Classification: 4.5 ($N_L$=6); MTOP NSP: 3.83 ($N_L$=6)
\end{itemize}

One can infer from the above numbers that behaviour of the model in our setup is sensitive to changes in random seed, model initialization and task at hand. We therefore stress that it is important to present results in our setup averaged across different seeds (as we have done in our work).

\end{document}